\documentclass[sigconf]{acmart}

\copyrightyear{2023} 
\acmYear{2023} 
\setcopyright{acmlicensed}\acmConference[GeoAI '23]{6th ACM SIGSPATIAL International Workshop on AI for Geographic Knowledge Discovery}{November 13, 2023}{Hamburg, Germany}
\acmBooktitle{6th ACM SIGSPATIAL International Workshop on AI for Geographic Knowledge Discovery (GeoAI '23), November 13, 2023, Hamburg, Germany}
\acmPrice{15.00}
\acmDOI{10.1145/3615886.3627749}
\acmISBN{979-8-4007-0348-5/23/11}

\usepackage[ruled,vlined]{algorithm2e}

\graphicspath{ {./figures/} }
\usepackage{comment}
\usepackage{amsmath,amsfonts}

\usepackage{adjustbox}
\usepackage[flushleft]{threeparttable}
\usepackage[utf8]{inputenc}
\usepackage{diagbox} 
\usepackage{longtable}
\usepackage{booktabs}
\usepackage{soul}
\usepackage{flushend}

\usepackage{booktabs, caption, tabularx, makecell}

\begin{document}

\title{Tensor Dirichlet Process Multinomial Mixture Model with Graphs for Passenger Trajectory Clustering}


\author{Ziyue Li}
\authornote{Corresponding Author.}
\email{zlibn@wiso.uni-koeln.de}
\orcid{0000-0003-4983-9352}
\affiliation{%
  \institution{University of Cologne}
  \streetaddress{Albertus-Magnus-Platz}
  \city{Cologne}
  \state{NRW}
  \country{Germany}
  \postcode{50923}
}

\author{Hao Yan}
\email{hyan46@asu.edu}
\affiliation{%
 \institution{Arizona State University}
 \city{Tempe}
 \country{U.S.A}}

\author{Chen Zhang}
\email{zhangchen01@tsinghua.edu.cn}
\affiliation{%
 \institution{Tsinghua University}
 \streetaddress{602 Shunde building}
 \city{Beijing}
 \country{China}}

\author{Lijun Sun}
\email{lijun.sun@mcgill.ca}
\affiliation{%
  \institution{McGill University}
  \streetaddress{817 Sherbrooke St. W.
Macdonald ENG BLDG 278C}
  \city{Montreal}
  \state{QC H3A 0C3}
  \country{Canada}
}

\author{Wolfgang Ketter}
\email{ketter@wiso.uni-koeln.de}
\affiliation{%
  \institution{University of Cologne}
  \streetaddress{Albertus-Magnus-Platz}
  \city{Cologne}
  \state{NRW}
  \country{Germany}
  \postcode{50923}
}

\author{Fugee Tsung}
\email{season@ust.hk}
\affiliation{%
  \institution{The Hong Kong University of Science and Technology}
  \streetaddress{Univresity Road}
  \city{Kowloon}
  \country{Hong Kong SAR}
  \postcode{50923}}
\affiliation{%
  \institution{HKUST-GZ}
  \city{Guangzhou}
  \country{China}
}
\renewcommand{\shortauthors}{Li and Yan, et al.}

\begin{abstract}
Passenger clustering based on trajectory records is essential for transportation operators. 
However, existing methods cannot easily cluster the passengers due to the hierarchical structure of the passenger trip information, including multiple trips within each passenger and multi-dimensional information about each trip. Furthermore, existing approaches rely on an accurate specification of the clustering number to start.  Finally, existing methods do not consider spatial semantic graphs such as geographical proximity and functional similarity between the locations. In this paper, we propose a novel tensor Dirichlet Process Multinomial Mixture model with graphs, which can preserve the hierarchical structure of the multi-dimensional trip information and cluster them in a unified one-step manner with the ability to determine the number of clusters automatically. The spatial graphs are utilized in community detection to link the semantic neighbors. We further propose a tensor version of Collapsed Gibbs Sampling method with a minimum cluster size requirement. A case study based on Hong Kong metro passenger data is conducted to demonstrate the automatic process of cluster amount evolution and better cluster quality measured by within-cluster compactness and cross-cluster separateness. The code is available at \url{https://github.com/bonaldli/TensorDPMM-G}.
\end{abstract}




\maketitle

\section{Introduction}

\begin{figure}[t]
\centering
\includegraphics[width=\columnwidth]{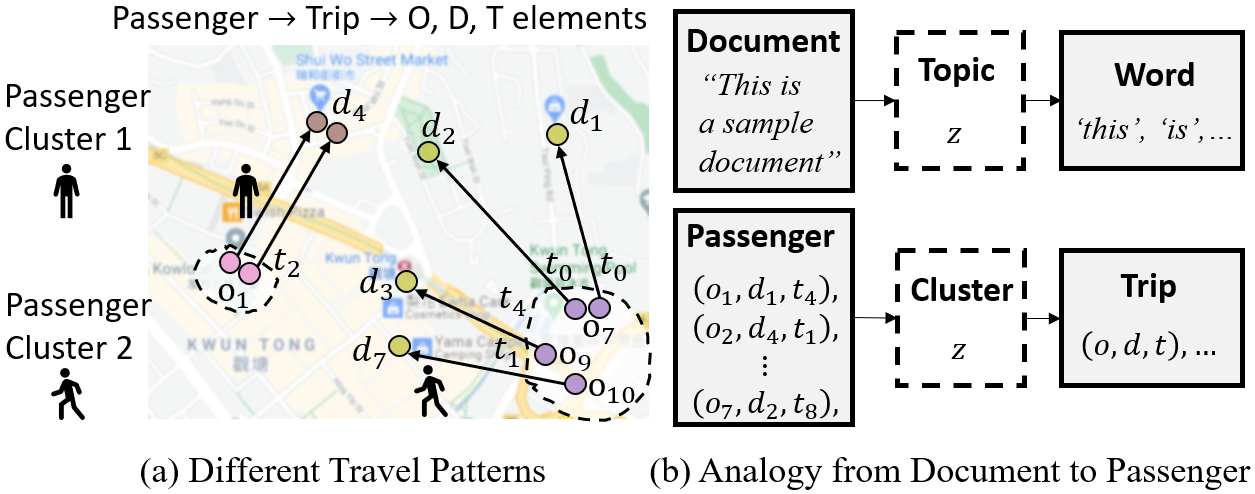}
\caption{(a) Passenger's hierarchical spatiotemporal trajectories; (b) Analogy from document, word to passenger, trip}
\label{analogy}
\end{figure}

Passenger trip data have a hierarchical structure, which we define formally as two layers from passengers to trips and trips to spatiotemporal components. Firstly, passengers have multiple trips with different amounts. Secondly, each trip contains abundant multi-dimensional spatiotemporal information,
i.e., that passenger $u$ departs from origin $o$ at time $t$ and arrives at destination $d$ at time $t'$, which is of significant research value to discover the useful insights about passenger clusters \cite{7891954, he2020classification, mao2022jointly}. This information could leverage various applications, such as prediction \cite{li2020long,lin2023dynamic,chen2023adaptive}, load inference\cite{jiang2023unified}, and personalization. This paper focuses on customization based on clustering.
For instance, as shown in Figure \ref{analogy}. (a), 
passenger cluster 2 has more trips overall, and their trips' spatiotemporal features are more diverse compared to cluster 1 who has a stable routine. Passenger clustering could further enable commercial applications for intelligent transportation systems, such as customized promotions and anomalous travel detection 
\cite{liu2017personalized, liu2019visual, li2022profile}.

Common methods for clustering passengers are based on temporal features such as time-series distance \cite{he2020classification}, active time \cite{briand2017analyzing}, and boarding time \cite{mohamed2016clustering}. However, these methods only consider the features in the temporal dimension yet ignore much information in spatial dimensions.

To consider all the spatiotemporal dimensions with a different number of trajectories is the first challenge. A few models are proposed \cite{cheng2020probabilistic, ziyue2021tensor} to firstly learn the hidden features from each dimension, such as frequent spatial and temporal features, co-occurrence, origin-destination pair \cite{yi2019machine}, and represent a passenger as a probabilistic representation over the extracted features; In the second step, passengers are clustered based on their representations using traditional distance-based methods. 
For example, as one of our prior works, Graph-Regularized Tensor LDA (3d-LDA-G) demonstrates the advantages of applying the traditional topic model in multi-clustering of origin, destination, time, and passenger \citep{li2022individualized}. \textcolor{black}{However, these two-step methods suffer from high computational costs and sub-optimal accuracy. For instance, the complexity of calculating the passenger representation's distance is $\mathcal{O}(M^2)$, which is not an optimal choice with millions of the passenger number $M$.} \textcolor{black}{Besides, another drawback of 3d-LDA-G lies in the basic assumption of LDA-based models, which is: each word has an independent topic assignment. This assumption renders a complex model and a long training time.}

To cluster passengers in a one-step manner, researchers have proposed generative models such as unigram multinomial mixture \cite{mohamed2016clustering}, Gaussian multinomial mixture \cite{briand2017analyzing}, Dirichlet multinomial mixture \cite{ziyue2021tensor}. Their first multinomial layer directly decides a passenger's cluster with the number of clusters as $K$. 
However, the second challenge lies exactly in the lack of prior knowledge to specify $K$. Take the Hong Kong metro system as an example, there are two million passengers daily, and Hong Kong also welcomes thousands of new arrivals flowing into the metro system, rendering it rather difficult for the system operators to specify the number of clusters.

Moreover, spatial correlations are believed to affect passenger mobility patterns, which are ignored by most clustering methods. Intuitively, if two locations are geographically adjacent or functionally similar, they may attract the same passengers \cite{geng2019spatiotemporal, li2020tensor, ziyue2021tensor, wang2023correlated}.

This encourages us to develop a method that can fulfill the following three desirable qualities: (1) able to handle hierarchical multi-dimensional data, (2) achieve one-step dynamic clustering with automatic $K$ determination, (3) consider the spatial correlations. To this end, we innovatively propose a \textbf{Tensor} \textbf{D}irichlet \textbf{P}rocess \textbf{M}ultinomial \textbf{M}ixture model with \textbf{G}raphs (Tensor-DPMM-G). To the best of our knowledge, we are the first work that fulfills all three qualities.

The greatest novelty of our proposed method lies in its interdisciplinary application of the traditional short text clustering model \cite{yin2016model} into the passenger clustering problem. (1) To preserve the innate hierarchical spatiotemporal multi-dimensional data, a tensor structure \cite{kolda2009tensor} is adopted. 
As shown in Figure \ref{analogy}. (b), each trip is represented as a three-dimensional word $\boldsymbol{w} = (w^O, w^D, w^T)$; A passenger with several trips, thus ``a small bag of words'', is viewed as a three-dimensional text $\boldsymbol{\mathbf{d}_u} \in \mathbb{R}^{O \times D \times T}$; The cluster assignment $z$ is the latent topic layer. (2) The base model, the Dirichlet Multinomial Mixture model \cite{yin2014dirichlet}, directly samples a membership for each passenger, achieving one-step clustering; The Dirichlet process \cite{gershman2012tutorial, li2019tutorial} 
allows each passenger to either choose an existing cluster or create a new cluster in a dynamic, flexible, non-parametric, and emerging-with-data manner, saving us from the burden to fix the number of clusters. (3) The spatial correlations are formulated as semantic graphs, i.e., a geographical proximity graph and functional similarity graph, to detect the hidden communities for origin and destination.

In summary, our technical contributions are as follows: 
\begin{itemize}
   
    \item We propose a tensor DPMM model to cluster passenger multi-dimensional trajectory data in a one-step manner with automatic $K$ determination; 
    
    \item 
    We formulate two spatial semantic graphs and adopt a community detection method to link graph neighbors;
    
    \item Besides, we also propose a tensor collapsed Gibbs sampling algorithm with a minimal cluster size requirement and a relocating step; 
    
    \item An experiment is conducted in real metro passenger trajectory data. We demonstrate that the cluster amount is determined automatically, and the learned clusters are four times more meaningful than benchmark methods.
    
\end{itemize}


The rest of this paper is organized as follows. We first give our model preliminaries; Then, we propose the Tensor-DPMM-G model, followed by an estimation algorithm; A real case study is conducted in the experiment. Finally, we give the conclusion and future work.

\section{Preliminary}

\textbf{Dirichlet Multinomial Mixture (DMM)} \cite{yin2014dirichlet}: Known as a short text model, this model has two major assumptions: (1) the text is short (often around $10^2$), and (2) the whole text only has one topic. 

A document $\mathbf{d}$ is generated by a mixture model: the cluster assignment $k$ for $\mathbf{d}$ is selected with weights $P(z=k)$; $\mathbf{d}$ is generated from distribution $p(\mathbf{d} | z=k)$. The likelihood of $\mathbf{d}$ could be represented as a mixture over all the $K$ clusters: $P(\mathbf{d}) = \sum_{k=1}^{K} P(\mathbf{d} | z=k) P(z=k)$. Once a document's cluster is assigned as $k$, all the words from this document are all generated independently using this cluster $k$ as the topic: $P(\mathbf{d} | z=k) = \prod_{ w \in \mathbf{d}} P(w | z=k)$. Then the document likelihood could be written as follows:
\begin{equation}
    P(\mathbf{d}) =  \sum_{k=1}^{K}  P(z=k) \prod_{ w \in \mathbf{d}} P(w | z=k)
\end{equation}

We also observe the same patterns in passenger data: (1) 84\% of the passengers only have one or two trips per day, so the average number of trips of a passenger over a whole year is around $10^2$; (2) Those passengers usually have only one travel purpose (school, work, etc.) \cite{zhao2020discovering}, so all the trips from the same passenger should share one topic once the passenger cluster is assigned. 

The DMM model could achieve direct one-step clustering. However, an automatic $K$ determination is desired. 

\textbf{Dirichlet Process (DP)} \cite{li2019tutorial}: DP allows $K$ to represent a varying number of clusters, whether finite or infinite. A fixed $K$ is no longer required. The DP is often explained with a metaphor called the Chinese Restaurant Process (CRP): each data instance is a customer, and each cluster is a round table. We interchangeably use the word ``cluster'' and ``table''.

When a new customer $u$ arrives, he/she can either choose an existing non-empty table $\mathbf{z}_u=k$ with probability:
\begin{equation} \label{eq_prel_choosek}
    P(z_u = k | \mathbf{z}_{-u}, \alpha) = \frac{m_{k, -u}}{M-1+\alpha},
\end{equation}

or choose a new empty table $\mathbf{z}_u=K+1$ with probability:
\begin{equation} \label{eq_prel_choosenew}
    P(z_u = K+1| \mathbf{z}_{-u}, \alpha) = \frac{\alpha}{M-1+\alpha}.
\end{equation}
where $K$ is the number of tables, $M$ is the total amount of passengers, $\alpha$ is a prior parameter, $\mathbf{z}_{-u}$ is the cluster assignment vector without the $u$-th passenger,  and $m_{k, -u}$ is the number of passengers in the table $k$ except the passenger $u$.
However, the DMM and DP are not directly applicable to multi-dimensional data like the passenger trajectory data.

\begin{figure}[t]
\centering
\includegraphics[width=0.95\columnwidth]{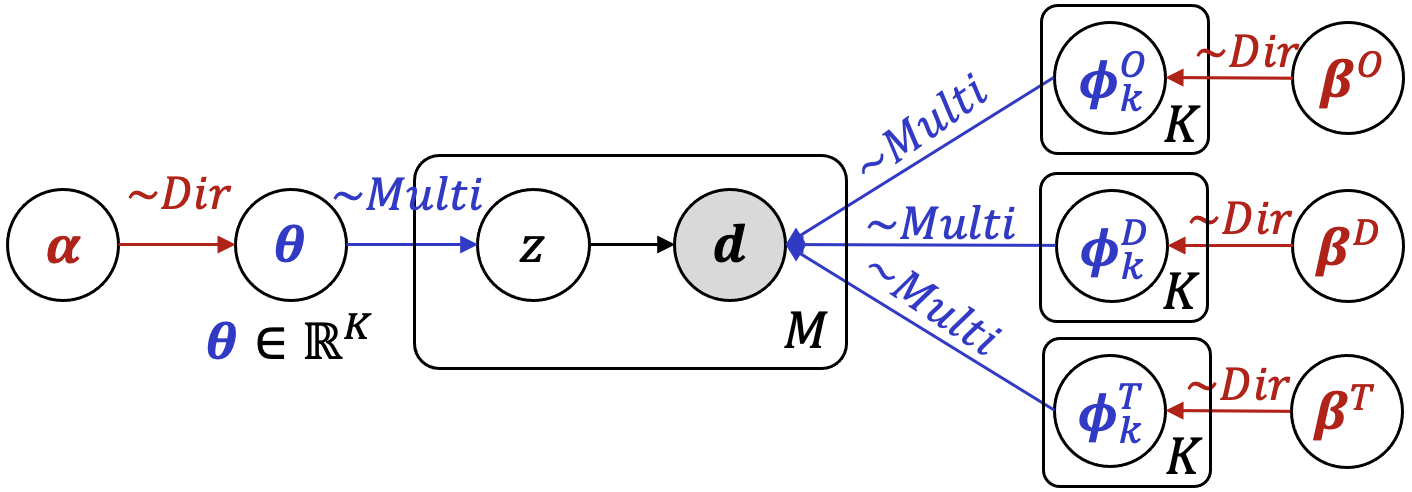}
\caption{Generative Process for TensorDMM Model}
\label{fig_gen_tensordmm}
\end{figure}

\section{Formulation}

\subsection{Notation}

Through out this exposition, scalars are denoted in italics, e.g. $n$; vectors by lowercase letters in boldface, e.g. $\mathbf{u}$; and matrices by uppercase boldface letters, e.g. $\mathbf{U}$.

\subsection{Multi-dimensional Word and Document}

A trip is defined as a three-dimensional word $\boldsymbol{w} = (w^O, w^D, w^T)$, indicating a trajectory starting from origin $w^O$ at time $w^T$ and arriving at destination $w^D$. $V^O, V^D, V^T$ are the vocabulary size for $w^O, w^D, w^T$, respectively. The passenger $u$ who has traveled $N_u$ trips could be viewed as a short document: $ \mathbf{d}_u = \{\boldsymbol{w}_1, \dots, \boldsymbol{w}_i, \dots, \boldsymbol{w}_{N_u}\}$, where $\boldsymbol{w}_i$ is the $i$-th trip of the passenger $u$. The entire data are $\mathbf{D} = \{ \mathbf{D}_u\}_{u=1}^{M}$. 
The concepts of word and document are interchangeable with trip and passenger respectively. 

Similarly, the passenger $\mathbf{d}_u$ is generated from a mixture model. Once a passenger's cluster is assigned as $k$, all his/her trips along dimensions of origin, destination, and time (ODT) are all generated independently using this cluster $k$ as the topic: $ P(\mathbf{d}_u | z=k) = \prod_{i=1}^{N_u} P(w^O_i | z=k) P(w^D_i | z=k) P(w^T_i | z=k)$. Therefore, the passenger likelihood could eventually be represented as follows: 

\begin{equation}
\begin{split}
    & P(\mathbf{d}_u) = \sum_{k} P(z=k) \times\\
    & \prod_i^{N_u} P(w^O_i | z=k) P(w^D_i | z=k) P(w^T_i | z=k).
\end{split}
\label{eq_cp}
\end{equation}

We denote $P(z=k) = \theta_k$ and $\sum_{k} \theta_k=1$, and $ P(w^O_i = o | z=k) = \phi_{ko}^{O}, P(w^D_i = d | z=k) = \phi_{kd}^{D}, P(w^T_i = t | z=k) = \phi_{kt}^{T}$. 

\subsection{Tensor Dirichlet Multinomial Mixture}

We will start with the basic version of the proposed model: a Tensor Dirichlet Multinomial Mixture (TensorDMM) model. 
Later we mainly emphasize the significant extensions based on this version. 

TensorDMM is designed as a generative model with multiple layers, which is suitable for the hierarchical trajectory data. The first layer generates a cluster $z_u$ via a multinomial distribution given passenger $u$ and the parameter $\boldsymbol{\theta}$. 
The second layer generates the word elements $w^O, w^D, w^T$ via three multinomial distributions given the sampled cluster $z_u$ and the parameters $ \boldsymbol{\phi}_k^O, \boldsymbol{\phi}_k^D$, $\boldsymbol{\phi}_k^T$. 
The latent parameters $\boldsymbol{\theta}, \boldsymbol{\phi}_k^O, \boldsymbol{\phi}_k^D$, and $\boldsymbol{\phi}_k^T$ are drawn from Dirichlet distributions with prior parameters $\boldsymbol{\alpha}, \boldsymbol{\beta}^O, \boldsymbol{\beta}^D$, and $\boldsymbol{\beta}^T$, respectively. 

\begin{equation} \label{eq_gen_tensordmm}
\begin{split}
    \boldsymbol{\theta} &\sim \text{Dir}(\boldsymbol{\alpha})\\
    z_u &\sim \text{Multi}(\boldsymbol{\theta}), \forall u \in M\\
    \boldsymbol{\phi}_k^O \sim \text{Dir}(\boldsymbol{\beta}^O), \boldsymbol{\phi}_k^D &\sim \text{Dir}(\boldsymbol{\beta}^D), \boldsymbol{\phi}_k^T \sim \text{Dir}(\boldsymbol{\beta}^T), \forall k \in K\\
    w^O \sim \text{Multi}(\boldsymbol{\phi}_{z_u}^O), w^D &\sim \text{Multi}(\boldsymbol{\phi}_{z_u}^D), w^T \sim \text{Multi}(\boldsymbol{\phi}_{z_u}^T)
\end{split}
\end{equation}

The whole generative process is shown in Figure \ref{fig_gen_tensordmm}.

\subsection{Tensor Dirichlet Process Multinomial Mixture Model with Graphs}

\subsubsection{Generative Process:}
The TensorDMM model is formulated in a parametric form, with the number of clusters fixed as $K$. The proposed Tensor Dirichlet Process Multinomial Mixture (TensorDPMM) model instead relaxes the restriction and can be considered as an infinite extension of the TensorDMM. The TensorDPMM model replaces Dirichlet prior $\text{Dir}(\boldsymbol{\alpha})$ with a stick-breaking construction \cite{teh2007stick}, $\boldsymbol{\theta} \sim \text{GEM}(1, \boldsymbol{\alpha})$. As a result, the number of clusters is emerging with the data dynamically.

\begin{equation} \label{eq_gen_tensordmm}
\begin{split}
    \boldsymbol{\theta} &\sim \text{GEM}(1, \boldsymbol{\alpha})\\
    z_u &\sim \text{Multi}(\boldsymbol{\theta}), \forall u \in M \\
    \boldsymbol{\phi}_k^O \sim \text{Dir}(\boldsymbol{\beta}^O), \boldsymbol{\phi}_k^D &\sim \text{Dir}(\boldsymbol{\beta}^D), \boldsymbol{\phi}_k^T \sim \text{Dir}(\boldsymbol{\beta}^T), \forall k \in \infty\\
    w^O \sim \text{Multi}(\boldsymbol{\phi}_{z_u}^O), w^D &\sim \text{Multi}(\boldsymbol{\phi}_{z_u}^D), w^T \sim \text{Multi}(\boldsymbol{\phi}_{z_u}^T)
\end{split}
\end{equation}

\begin{figure}[t]
\centering
\includegraphics[width=0.95\columnwidth]{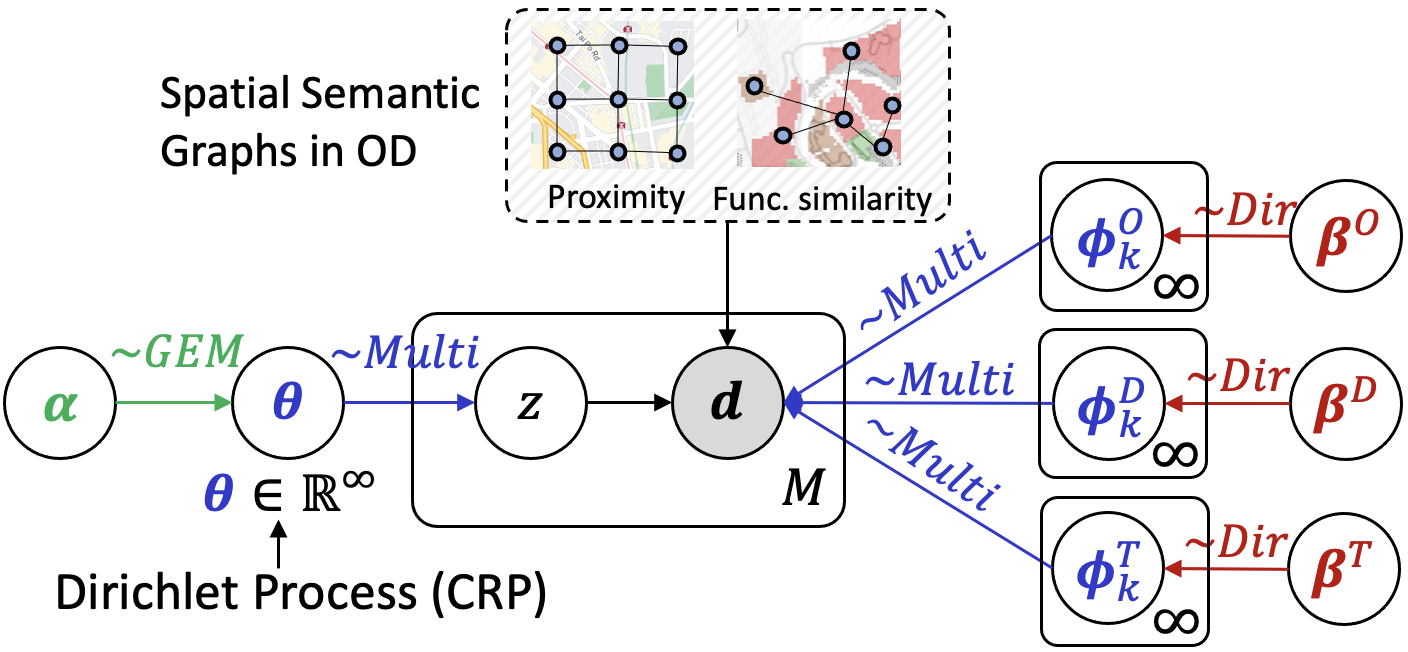}
\caption{Generative Process for TensorDPMM-G Model}
\label{fig_gen_tensordpmm}
\end{figure}

\subsubsection{Choose A Table - Dirichlet Process:}
In the TensorDMM model, the Multinomial parameter to choose a cluster has a fixed dimension $\boldsymbol{\theta} \in \mathbb{R}^K$; However, in the TensorDPMM model, this parameter grows with the data $\boldsymbol{\theta} \in \mathbb{R}^\infty$, making the probability of a passenger choosing a cluster also dynamically evolve with data. Thus, the core of the Dirichlet process 
is to define the following two probabilities: (1) choosing an existing table and (2) choosing a new table.

\textbf{Choosing An Existing Table:} The probability of the passenger $\mathbf{d}_{u}$ choosing an existing cluster $z$ given the other documents and their clusters is:

\begin{equation}\label{eq_prob_choosez}
\begin{split}
    &P(z_u = z | \mathbf{z}_{-u}, \mathbf{D}, \mathbf{\alpha}, \mathbf{\beta}^O, \mathbf{\beta}^D, \mathbf{\beta}^T) \\
    &\propto P(z_u = z | \mathbf{z}_{-u}, \mathbf{D}_{-u}, \mathbf{\alpha}, \beta^O, \beta^D, \beta^T) \times\\ 
    &{\,\,\,\,\,\,\,}P(\mathbf{d}_{u} | z_u = z, \mathbf{z}_{-u}, \mathbf{D}_{-u}, \mathbf{\alpha}, \beta^O, \beta^D, \beta^T)\\
    &\propto P(z_u = z | \mathbf{z}_{-u}, \mathbf{\alpha}) \times P(\mathbf{d}_{u} | z_u = z, \mathbf{D}_{z, -u}, \beta^O, \beta^D, \beta^T)
\end{split}
\end{equation}
\noindent where $\mathbf{D}_{-u}$ ($\mathbf{D}_{z, -u}$) is all the passengers (in the table $z$) except the $u$-th one.  
The first and second terms in Eq. (\ref{eq_prob_choosez}) are derived prudently with details in the technical appendix.

The first term is as follows:

\begin{equation} \label{eq_choosez_term1}
    P(z_u = z | \mathbf{z}_{-u}, \mathbf{\alpha}) = \frac{m_{z,-u} + \frac{\alpha}{K}}{M-1+\alpha}
\end{equation}
When $K \gg \alpha$, Eq. (\ref{eq_choosez_term1}) reduces to Eq. (\ref{eq_prel_choosek})

The second term is as follows:
\begin{equation} \label{eq_choosez_term2}
\begin{split}
    & P(\mathbf{d}_{u} | z_u = z, \mathbf{D}_{z, -u}, \beta^O, \beta^D, \beta^T) \\
    & \propto \prod_{ODT} \left\{ \frac{\prod_{w \in \mathbf{d}_u} \prod_{j=1}^{N^{w}_d} (n_{z,-u}^{w} + \mathbf{\beta} +j - 1) }{\prod_{i=1}^{N_u}(n_{z,-u} + V \beta + i - 1)} \right\}
\end{split}
\end{equation}

Eq. (\ref{eq_choosez_term2}) shows that, in a high-dimensional case, the second term is the decomposition result over all dimensions ODT.

Finally, the probability of passenger choosing the existing cluster $z$ given the other documents and their clusters is:
\begin{equation}\label{eq_prob_choosez_final}
\begin{split}
    & P_z = P(z_u = z | \mathbf{z}_{-u}, \mathbf{D}, \mathbf{\alpha}, \mathbf{\beta}^O, \mathbf{\beta}^D, \mathbf{\beta}^T) \\
    & \propto \frac{m_{z,-u} + \frac{\alpha}{K}}{M-1+\alpha} \prod_{ODT} \left\{ \frac{\prod_{w \in \mathbf{d}_u} \prod_{j=1}^{N^{w}_d} (n_{z,-u}^{w} + \mathbf{\beta} +j - 1) }{\prod_{i=1}^{N_u}(n_{z,-u} + V \beta + i - 1)} \right\}
\end{split}
\end{equation}

In Eq. (\ref{eq_prob_choosez_final}), the first term increases with a larger $m_{z,-u}$, which reflects the Rule-1 of ``choosing a table with more people''; The second term instead measures the ratio of the word in the passenger $u$ over the word in the cluster $z$, and it increases when cluster $z$ has more same words with the passenger $u$, which reflects the Rule-2 of ``choosing a table with more commonalities'' \cite{yin2016model}. 

\textbf{Choosing A New Table:} 
To choose a new table, the number of clusters increases to $K+1$. Similarly, the corresponding probability of passenger $u$ choosing the new cluster is:

\begin{equation} \label{eq_prob_choose_new}
\begin{split}
    &P(z_u = K+1 | \mathbf{z}_{-u}, \mathbf{D}, \mathbf{\alpha}, \beta^O, \beta^D, \beta^T) \\
    &\propto P(z_u = K+1 | \mathbf{z}_{-u}, \mathbf{\alpha}) \times \\
    &{\,\,\,\,\,\,\,} P(\mathbf{d}_{u} | z_u = K+1, \mathbf{D}_{z, -u}, \beta^O, \beta^D, \beta^T)
\end{split}
\end{equation}

The first term is derived as follows:

\begin{equation} \label{eq_prob_choose_new_term1}
\begin{split}
    & P(z_u = K+1 | \mathbf{z}_{-u}, \mathbf{\alpha}) 
    = \frac{\alpha}{M-1+\alpha}
\end{split}
\end{equation}

The second term is derived as follows:

\begin{equation} \label{eq_prob_choose_new_term2}
\begin{split}
    & P(\mathbf{d}_{u} | z_u = K+1, \mathbf{D}_{z, -u}, \beta^O, \beta^D, \beta^T)\\
    & \propto \prod_{ODT} \left\{ \frac{\prod_{w \in \mathbf{d}_u} \prod_{j=1}^{N_u^{w}} (\beta + j -1)}{\prod_{i=1}^{N_u} (V \beta + i - 1)} \right\}
\end{split}
\end{equation}

Therefore, the final probability of passenger $u$ choosing a new cluster $K+1$ is:

\begin{equation} \label{eq_prob_choose_new_final}
\begin{split}
    &P_{K+1} = P(z_u = K+1 | \mathbf{z}_{-u}, \mathbf{D}, \mathbf{\alpha}, \beta^O, \beta^D, \beta^T) \\
    &\propto \frac{\alpha}{M-1+\alpha} \prod_{ODT} \left\{ \frac{\prod_{w \in \mathbf{d}_u} \prod_{j=1}^{N_u^{w}} (\beta + j -1)}{\prod_{i=1}^{N_u} (V \beta + i - 1)} \right\}
\end{split}
\end{equation}

In Eq. (\ref{eq_prob_choose_new_final}), the first term still follows Rule-1
, where $\alpha$ is the pseudo number of members in the new cluster. Therefore with a larger $\alpha$, passenger tends to choose a new cluster; The second term still follows Rule-2
, where $\beta$ is the pseudo frequencies of each word in the new cluster.

\begin{figure}[t]
\centering
\includegraphics[width=0.95\columnwidth]{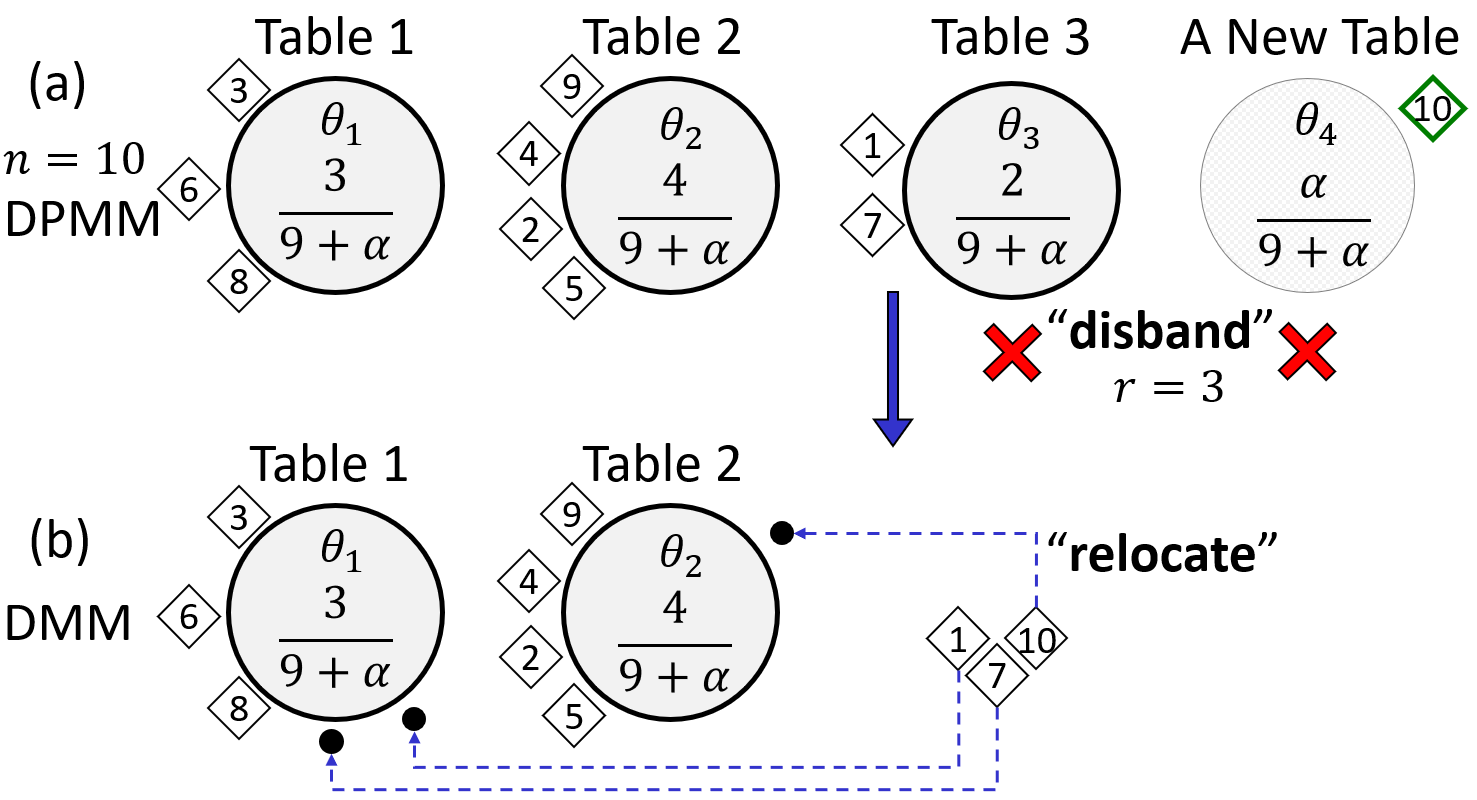}
\caption{
(a) An example for a new 10-th passenger choosing a table in a DPMM manner; (b) Then disbanding tables with $m_z<r$ and relocating members 1,7,10 in a DMM manner.}
\label{fig_disband_relocate}
\end{figure}

A toy example for the $10$-th passenger to choose an existing table or a new table is shown in Figure \ref{fig_disband_relocate}. (a).

\subsubsection{Spatial Semantic Graph:}

The essence of traditional short text clustering models is to mechanically put documents that have the exact same words into the same cluster. However, in our spatiotemporal trajectory data, spatial semantic graphs are observed in origin and destination dimensions \cite{geng2019spatiotemporal, li2020tensor}. This motivates us that, for instance in origin dimension, although $o_1$ and $o_2$ are different stations, if they are graph neighbors, passengers containing $o_1$ and passengers containing $o_2$ can be in the same cluster. 
Graphs are usually incorporated into topic models as graph Laplacian regularization \cite{mei2008topic, ziyue2021tensor}, but this will break the conjugate of the generative model. 

In this section, we formulate two spatial graphs, namely the geographical proximity graph and the functional similarity graph. 
Then we conduct a graph community detection \cite{kim2015community} to identify the communities in graphs, and replace the original word $w^O, w^D$ with the detected spatial communities $s \in S$, where $S$ is the total amount of detected communities. The rest of the model remains the same.



\textbf{Geographical Proximity Graph:} The distance from station $i$ to $j$ is calculated. The proximity graph is formulated as an "$h$-hop" binary graph: if the distance between station $i$ and $j$ is less than $h$-hops, two stations are adjacent. 

\begin{equation} \label{eq_gadj}
  \{\mathbf{G}_{adj}\}_{i,j}=\begin{cases}
              1, &  \text{hop distance}_{i,j} <= h\\
              0, &  \text{hop distance}_{i,j} > h\\
            \end{cases}
\end{equation}

\textbf{Functional Similarity Graph:} The functionality of a station can be characterized by the surrounding Point of Interest (POI), indicating the amount of the corresponding service (school, shopping malls, office buildings, etc.) nearby the station. 
Station $i$ and $j$ have similar functionality if the cosine similarity of their POIs is larger than a threshold $\gamma$:

\begin{equation} \label{eq_gpoi}
  \{\mathbf{G}_{poi}\}_{i,j}=\begin{cases}
              1, & \text{cos\_sim} \left(\textbf{POI}_{i},  {\textbf{POI}_{j}} \right) >= \gamma\\
              0, &  \text{otherwise}\\
            \end{cases}
\end{equation}

This graph community detection not only merges two same semantic locations together but also functions as a dimension reduction for OD. This technique is critical when dealing with a large transport network whose locations $V^O$ and $V^D$ (e.g., $\approx 10^3$) could be reduced to $S$ ($\approx 10^1$). 

The proposed model is summarized in Figure \ref{fig_gen_tensordpmm}.

\section{Algorithm}

This section provides a tensor version of the collapsed Gibbs sampling method to learn the passenger cluster assignment \cite{griffiths2004finding, porteous2008fast}. 
Considering the operational cost in practice, the size of clusters should not be too small since it is too expensive to provide a customized operation for a niche group. Therefore, we introduce a minimal cluster size requirement to control the cluster resolution, denoted as $r$. Taking the metaphor of a Chinese restaurant, we require each table to seat at least $r$ people.

\begin{algorithm}[t]
\SetAlgoVlined
\textbf{Input}: $\{\mathbf{d}_u\}_{u=1}^{M}, r, \alpha, {\beta}^{O}, {\beta}^{D}, {\beta}^{T}, max\_iter$\\
\textbf{Output}: $K, z_u, \forall u$\\
\textbf{Phase 1: Initialization}:\\
$K = K_0 (= 1), m_z, n_z, n_z^{w^O}, n_z^{w^D}, n_z^{w^T} =0$\\
\For{$u = 1$ to $M$}{
    Sample a cluster for $u$: $z_u \sim \text{Multi}(\frac{1}{K})$\\
    $m_z = m_z + 1, n_z = n_z + N_u$\\
    \For{$i = 1$ to $N_u$}{
        $n_z^{w^O} = n_z^{w^O} + N_u^{w^O}$, same for D,T}
    }
\textbf{Phase 2: Gibbs Sampling:}\\
\For{$iter = 1$ to $max\_iter$}{
    \For{$u = 1$ to $M$}{
        \textbf{Phase 2.1: ``Kick-out'': }\\
        $m_z = m_z - 1, n_z = n_z - N_u$\\
        \For{$i = 1$ to $N_u$}{
            $n_z^{w^O} = n_z^{w^O} - N_u^{w^O}$, same for D,T
        }
        \If{$n_z ==0$}{
            $K=K-1$
        }
        \textbf{Phase 2.2: ``Choose A Table'': }\\
        $z_u \sim \{P_z\}_{z=1}^{K}, P_{K+1}$ in Eq.(\ref{eq_prob_choosez_final}) for an existing table and Eq.(\ref{eq_prob_choose_new_final}) for a new table\\
        \textbf{Phase 2.3: ``Merge to the Selected Table'': }\\
        \If{$z = K+1$}{
            $K = K+1$
        }
        $m_z = m_z + 1, n_z = n_z + N_u$\\
        \For{$i = 1$ to $N_u$}{
            $n_z^{w^O} = n_z^{w^O} + N_u^{w^O}$, same for D,T
        }
    }
}
\textbf{Phase 3: ``Disband Small Tables and Relocate'': }\\
\For{$\forall z^* \in \{ z^* | m_{z^*} < r \}$}{
    \textbf{Phase 3.1: ``Disband'': }\\
    $K=K-1 $, delete $m_{z^*}, n_{z^*}, n_{z^*}^{w^O}, n_{z^*}^{w^D}, n_{z^*}^{w^T}$\\ 
    record $u^* \in \{ u^* | z_{u^*} = z^*\}$ \\}
\For{$u^*=1$ to $M^*$}{
    \textbf{Phase 3.2: ``Relocate to a remaining table'': }\\
    $z_u \sim \text{normalize}(\{P_z\}_{z=1}^{K})$ in Eq. (\ref{eq_prob_choosez_final})\\
    \textbf{Phase 3.3: ``Merge'':}  same as Phase 2.3\\
    }

\caption{Tensor $r$-Collapsed Gibbs Sampling}
\label{algo1}
\end{algorithm}

As shown in Algorithm \ref{algo1}, the proposed Tensor $r$-Collapsed Gibbs Sampling algorithm updates the parameters $m_z \in \mathbb{R}^K$ (number of passengers in cluster $z$), $n_z \in \mathbb{R}^K$ (number of words in cluster $z$), $n_z^{w^O} \in \mathbb{R}^{KV^O}$ (number of occurrences of word $w^O$ in cluster $z$), $n_z^{w^D}$, and $n_z^{w^T}$.

\begin{itemize}
    \item Phase 1 - Initialization: to assign all passengers an initial cluster; 
    \item Phase 2 - Gibbs Sampling: with the standard three steps of ``kick-out'' from the initial cluster, ``choose a table'' based on the two probabilities in Eq. (\ref{eq_prob_choosez_final}) and (\ref{eq_prob_choose_new_final}), and ``merge to the selected table''; The main difference is that word-level operations are conducted for ODT iteratively;
    \item Phase 3 - ``Disband Small Tables and Relocate'': At the end of Phase 2, we will check whether the formed tables fulfill the minimal table size requirement. As illustrated in Figure \ref{fig_disband_relocate}. (b), we will disband the tables with $m_{z^*} < r$ and relocates all their members $u^* = 1 \dots M^*$ ($M^*$ is the total population that needs to relocate) into the remaining tables as shown in Phase 3.2: It is worth mentioning that in this relocating step, members are not given a chance to choose a new table, so this relocating step reduces to a DMM model without $P_{K+1}$.
\end{itemize}

The computational complexity of each iteration is $\mathcal{O} \left( n_{dim}MNK \right)$, where $n_{dim}$ is the number of dimensions in the data, i.e., $n_{dim} = 3$ in our ODT data.

\section{Experiment}

In this section, we will implement the proposed model into Hong Kong metro passenger trajectory data.

\textbf{Dataset:} The trajectory data are chosen from 50,000 randomly picked passengers during the period of 01-Jan-2017 to 31-Mar-2017. The information of entry station $w^O$, entry time (in hour) $w^T$, and exit station $w^D$ are kept for analysis. The average length of each passenger is $\bar{N}  \approx 134$. The Hong Kong metro system had 98 stations in 2017 and operated in 24 hours. Thus the vocabulary size for origin, destination, and time is 98, 98 and 24.

\begin{figure}[t]
\centering
\includegraphics[width=0.9\columnwidth]{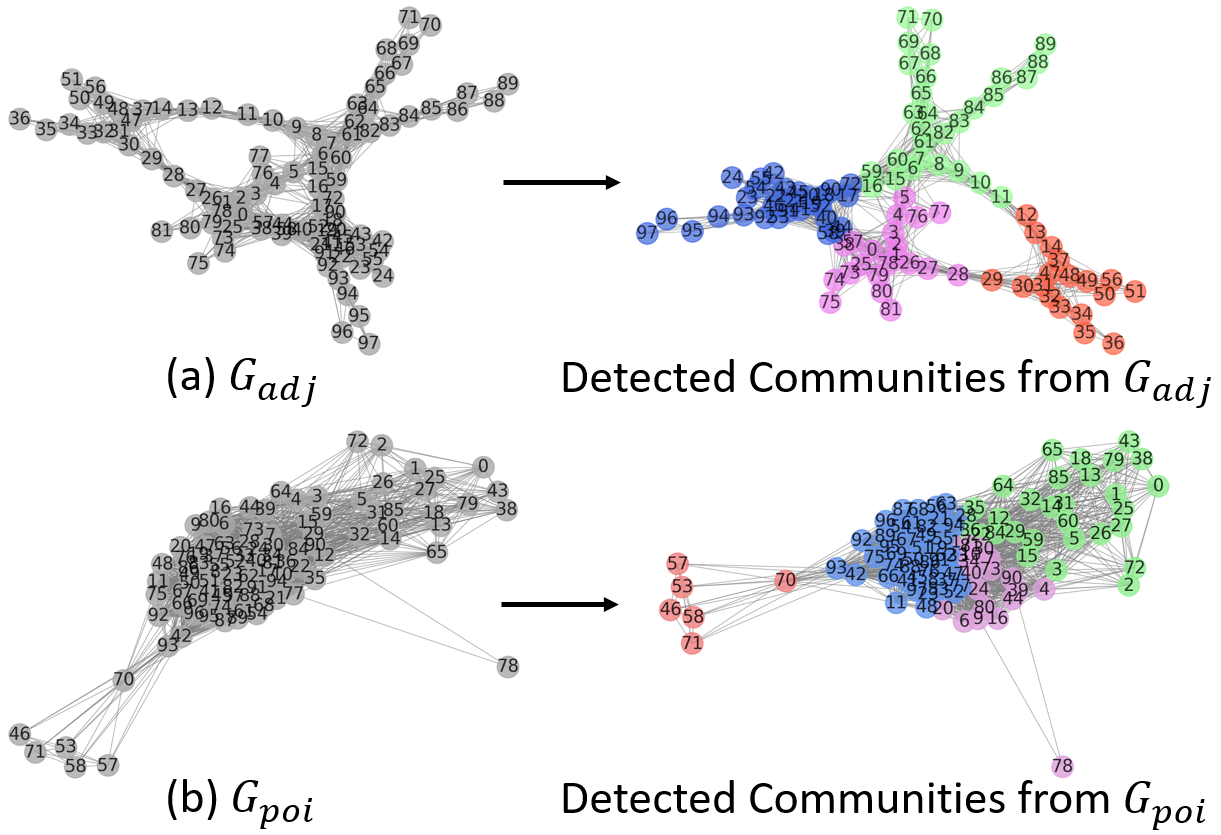}
\caption{Community Detection for (a) $\mathbf{G}_{adj}$, (b) $\mathbf{G}_{poi}$.}
\label{detected_community}
\end{figure}

\textbf{Graph Community Detection:}
As mentioned before, geographical proximity graph $\mathbf{G}_{adj}$ and functional similarity graph $\mathbf{G}_{poi}$ are observed in the dimension of OD. In Eq. (\ref{eq_gadj}) and (\ref{eq_gpoi}), we set $h=4$ and $\gamma = 0.7$, such that the communities for each graph are detected with maximized modularity using the Leiden algorithm \cite{traag2019louvain}. As shown in Figure \ref{detected_community}, four communities are detected in both graphs. Then each word component $w^O, w^D$ will be replaced by their belonged communities, with community amount as $4 \times 4$. After this pre-processing, the data will be fed into the proposed model.
%

\textbf{Evaluation Metrics:} Since our clustering is conducted on real data without labels, traditional evaluations such as the normalized mutual information are not applicable since they require labels as ground truth. We will use internal evaluation measures in Eq. (\ref{evaluation}) to evaluate the compactness, separateness, and overall quality of clusters \cite{liu2010understanding}.

Root-mean-square standard deviation (RMSSTD): It evaluates the within-cluster compactness. A smaller RMSSTD means a more compact cluster.

R-Squared (RS): It measures cross-cluster separateness. A bigger RS means more separate different clusters.

\textit{Calinski}-\textit{Harabasz} index (CH): It measures the two criteria above simultaneously by the formulation of $\frac{\text{separateness}}{\text{compactness}}$. 

The optimal value for RMSSTD and RS is the first “elbow” in their monotonically increasing or decreasing curves \cite{hassani2017using}.
The optimal value of CH is strictly the maximal.

\begin{equation}
\begin{split}
    & \text{RMSSTD} = \left( \frac{\sum_{k} \sum_{u \in C_k} \|\mathbf{d}_u- \mathbf{c}_k \|^2}{V^O V^D V^T \sum_{k}(m_k-1)} \right)^{1/2}\\
    & \text{RS} = \frac{\sum_{u \in M} \| \mathbf{d}_u - \mathbf{g}\|^2 - \sum_k \sum_{u \in C_k} \|\mathbf{d}_u- \mathbf{c}_k \|^2}{\sum_{u \in M} \| \mathbf{d}_u - \mathbf{g}\|^2}\\
    & \text{CH} = \frac{\sum_{k} \|\mathbf{c}_k - \mathbf{g} \|^2 / (K-1)}{\sum_k \sum_{u \in C_k} \|\mathbf{d}_u- \mathbf{c}_k \|^2 / (M - K)}
\label{evaluation}
\end{split}
\end{equation}
where $C_k$ is the $k$-th cluster with its center as $\mathbf{c}_k$, and $\mathbf{g}$ is the global center of dataset.

\textbf{Benchmark Methods:} We compare our method with the following benchmark methods.
\begin{itemize}
    \item \textbf{K-means}: It is one of the most common clustering methods. However, it could only handle one-dimensional data \cite{hartigan1979algorithm, 7891954}. 
    \item \textbf{Temporal Behaviour Clustering} (TBC): Clustering based on passengers' temporal patterns is widely used \cite{he2020classification, briand2017analyzing}. A mixture of unigrams \cite{mohamed2016clustering} model is used in this paper for comparison.
    \item \textbf{DMM}: This is a popular short-text model \cite{yin2014dirichlet}. However, it can only handle one-dimensional words. To apply it, we either only utilize one dimension (e.g., origin, denoted as DMM-o) or flatten the three-dimensional word $(w^O, w^D, w^T) = (o, d, t)$ into one-dimensional $w = odt$ (denoted as DMM-1d), which exponentials the vocabulary size to $V^O V^D V^T$.  
    \item \textbf{Three-dimensional Latent Dirichlet Allocation} (3d-LDA-G): It defines a generative process with a word-level topic along each dimension of ODT. Passengers then will be clustered based on their probabilistic distributions of the learned topics by distance-based methods \cite{cheng2020probabilistic}. Graphs can be added as Laplacian regularizations \cite{ziyue2021tensor}. 
    \item \textbf{TensorDMM}: This is the basic version of the proposed model, which clusters passengers using their multi-dimensional trip records in a one step manner.
    \item \textbf{TensorDPMM}: To validate the advantage of introducing spatial graphs, we also check the performance of the without-graph version.
\end{itemize}
\begin{table}[t]
\begin{threeparttable}
\setlength\tabcolsep{0pt} 
\begin{tabular*}{\columnwidth}{@{\extracolsep{\fill}} lccccccc}
\toprule
$r$ & 15 & 25 & 35 & 45 & 55 & 65 & 75\\
\midrule
$K$ & 78 & 48 & 30 & 23 & 21 & 15 & 14\\
RMSSTD & 11.65 & 11.87 & 12.03 & \textbf{12.25} & 12.31 & 12.44 & 12.53
\\
RS & 0.475 & 0.458 & 0.436 & \textbf{0.413} & 0.407 & 0.394 & 0.374\\
CH & 0.706 & 0.718 & 0.741 & \textbf{0.807} & 0.727 & 0.688 & 0.551\\
\bottomrule
\end{tabular*}
\caption{Resolution $r$ Sensitivity Analysis: Conducted with $\alpha, \beta^O, \beta^D=0.01, \beta^T = 0.042$}
\label{tab: r_search}
\end{threeparttable}
\end{table}

\begin{table}[t]
\begin{threeparttable}
\setlength\tabcolsep{0pt} 
\begin{tabular*}{\columnwidth}{@{\extracolsep{\fill}} lccccccc}
\toprule
$\alpha$ & 0.001 & 0.005 & 0.01 & 0.03 & 0.05 & 0.07 & 0.09\\
\midrule
$K$ & 26 & 25 & 23 & 25 & 21 & 26 & 25\\
RMSSTD & 12.34 & 12.34 & \textbf{12.25} & 12.47 & 12.43 & 12.47 & 12.35\\
RS & 0.404 & 0.403 & \textbf{0.413} & 0.392 & 0.395 & 0.392 & 0.404\\
CH & 0.704 & 0.729 & \textbf{0.807} & 0.684 & 0.738 & 0.684 & 0.756\\
\bottomrule
\end{tabular*}
\caption{Prior $\alpha$ Sensitivity Analysis: Conducted with $r =45, \beta^O, \beta^D=0.01, \beta^T = 0.042$.}
\label{tab: alpha_search}
\end{threeparttable}
\end{table}
\begin{figure}[t]
\centering
\includegraphics[width=0.85\columnwidth]{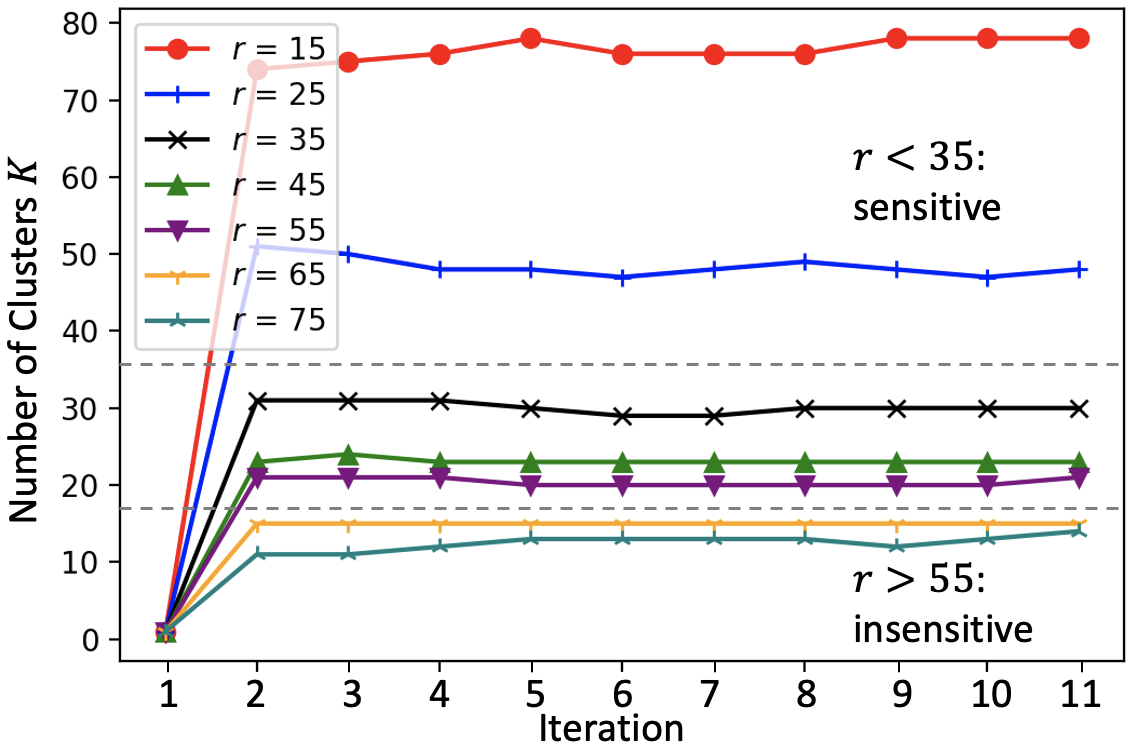}
\caption{$K$ Automatic evolution over iteration.}
\label{k_evolution}
\end{figure}

\begin{table*}[t]
\centering
\begin{tabularx}{0.65
\textwidth}{c|cccc|ccc}
\toprule
Methods & \textit{Tensor} & \textit{1-Step} & \textit{Auto-$K$} & \textit{Graph} & RMSSTD & RS & CH\\
\midrule
K-means &  $\times$ & $\checkmark$ & $\times$ & $\times$  & 30.47 & \underline{0.390} & 0.169\\
TBC & $\times$ & $\times$ & $\times$ & $\times$ & 23.59 & 1.491 & 0.215\\
DMM & $\times$ & $\checkmark$ & $\times$ & $\times$ & 23.64 & 1.319 & 0.277\\
\addlinespace
3d-LDA-G & $\checkmark$ & $\times$ & $\times$ & $\checkmark$ & 27.46 & 0.791 & 0.210\\
\addlinespace
TensorDMM & $\checkmark$ & $\checkmark$ & $\times$ & $\times$ & 23.61 & 1.450 & 0.280\\
TensorDPMM & $\checkmark$ & $\checkmark$ & $\checkmark$ & $\times$ & \underline{13.63} & 0.535 & \underline{0.676}\\
TensorDPMM-G & $\checkmark$ & $\checkmark$ & $\checkmark$ & $\checkmark$ & \textbf{12.25} & \textbf{0.413} & \textbf{0.807}\\
\bottomrule
\end{tabularx}
\caption{Internal Evaluation of Clustering Results from Different Methods}
\label{all_methods_comparison}
\end{table*}

\begin{figure*}[t]
\centering
\includegraphics[width=0.75\textwidth]{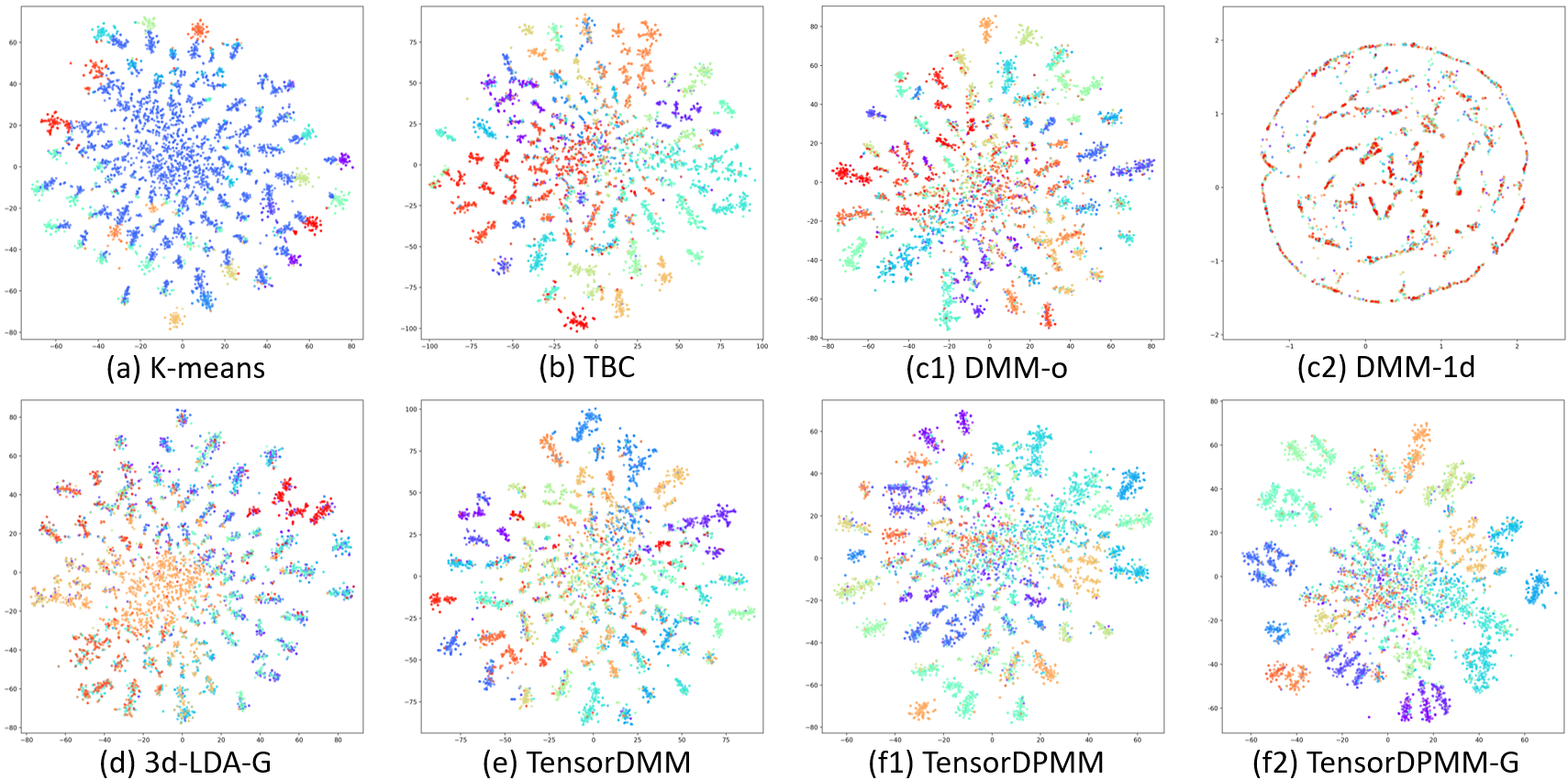}
\caption{Clustering Result Visualization via t-distributed Stochastic Neighbor Embedding \cite{van2008visualizing} 
} 
\label{cluster_visual}
\end{figure*}

We first compare all the methods by whether it is a tensor-based method to handle multi-dimensional data (\textit{Tensor} for short), cluster in a one-step manner (\textit{1-Step}), automatically determine $K$ (\textit{Auto-K}), and consider spatial semantic graphs (\textit{Graph}) in Table \ref{all_methods_comparison}. Only the proposed TensorDPMM-G full-version model ticks all the boxes.

\textbf{Best Parameter Searching and Sensitivity Analysis:}
The best set of parameters will be chosen via a grid search with the maximum CH index (or the first elbow of RMSSTD or RS value). In our case, $r=45, \alpha, \beta^O, \beta^D=0.01, \beta^T = 0.042$. Precisely, the minimum cluster size $r$ will directly affect the number of clusters: a larger $r$ will learn fewer clusters. $\alpha$ will also affect the number of clusters theoretically since a larger $\alpha$ means a passenger will be more likely to choose a new table. Therefore, here we will mainly demonstrate how $r$ and $\alpha$ will affect cluster learning as shown in Table \ref{tab: r_search} and Table \ref{tab: alpha_search}.

As shown in Table \ref{tab: r_search}, the maximum CH (same as the first elbow of RMSSTD and RS) happens at $r=45$, and $K$ is converged at $K=23$. The learned $K$ is more sensitive with $r$ when $r<35$ and less sensitive when $r>55$. The automatic and dynamic evolution of the number of clusters over iterations with different resolutions is shown in Figure \ref{k_evolution}.

As shown in Table \ref{tab: alpha_search}, $K$ is not strictly increasing with $\alpha$, and the main reason is that there are two rules mentioned before simultaneously affecting the choices of a cluster.

\textbf{Improved Performance:} As shown in Table \ref{all_methods_comparison}, the traditional one-dimensional methods  
(i.e., K-means, TBC, and DMM) have lousy performance overall since the rich multi-dimensional spatiotemporal information is abandoned. Specifically, (1) K-means learns a big blue cluster neighboring with other clusters as shown in Figure \ref{cluster_visual}. (a), which makes it have the largest RMSSTD yet relatively small RS; (2) For DMM-1d in Figure \ref{cluster_visual}. (c2), when the ODT data is flattened as one-dimension, the vocabulary space expands to $V^O V^D V^T \approx 2\times10^5$, making the data extremely sparse and hard to learn any meaningful clusters.

The tensor-based methods could achieve relatively better performance due to well-preserved ODT information. However, 3d-LDA-G is only to learn the local topics (clusters) in each word dimension, not a global cluster for passengers.

The proposed TensorDPMM-G 
achieves the highest 
CH value, four times higher than the benchmarks, proving that the proposed model can learn clusters with better qualities. Precisely, (1) the proposed model learns clusters with more within-cluster compactness (lowest RMSSTD), more cross-cluster separateness (highest RS), and the best overall quality (highest CH); (2) Futhermore, compared with our TensorDPMM without graphs, incorporating graphs further improves the clusters. As shown in Figure \ref{cluster_visual}. (f1) and (f2), after considering spatial graphs, the clusters in the same color (e.g., purple, dark blue, light green, etc.) are learned closer.

\section{Conclusion and Future Work}

This paper studied a passenger clustering problem based on the hierarchical, multi-dimensional, and spatiotemporal trip data. We proposed to use the tensor to preserve the spatiotemporal data and the Dirichlet process multinomial mixture to achieve the one-step clustering with automatic determination of the cluster number. We further incorporate two spatial semantic graphs into the data. In addition, we offered a tensor Gibbs sampling algorithm with a minimum cluster size requirement to learn the clusters. In the case study, we demonstrate the improved qualities of the learned clusters. 

For future research, we aim to embed graphs directly into the model's generative process as a unified model. Besides, we also plan to add trip duration as another dimension.

\bibliography{reference}
\bibliographystyle{ACM-Reference-Format}

\vfill


\end{document}